\documentclass[a4paper,conference]{IEEEtran}
\usepackage{graphicx} 
\usepackage{stfloats}
\usepackage{cite}
\usepackage{amsmath}
\usepackage{multirow}
\usepackage{tabularx}
\IEEEoverridecommandlockouts
\ifCLASSINFOpdf
\else
\fi
\newcommand{\proposed}{AFF-Net}
\newcommand{\etal}{\textit{et al.}}

\newcommand{\ie}{\textit{i.e.}}

\begin{document}
%
\title{Adaptive Feature Fusion Network for Gaze Tracking in Mobile Tablets}

\author{\IEEEauthorblockN{Yiwei Bao\textsuperscript{1} \qquad Yihua Cheng\textsuperscript{1} \qquad Yunfei Liu\textsuperscript{1} \qquad Feng Lu\textsuperscript{1,2,*}}
\IEEEauthorblockA{\textsuperscript{1}State Key Laboratory of Virtual Reality Technology and Systems, School of CSE, Beihang University, Beijing, China \\ \textsuperscript{2}Peng Cheng Laboratory, Shenzhen, China\\
\texttt{\{baoyiwei, yihua\_c, lyunfei, lufeng\}@buaa.edu.cn}}
}

%


\maketitle

\begin{abstract}

Recently, many multi-stream gaze estimation methods have been proposed.
They estimate gaze from eye and face appearances and achieve reasonable accuracy.
However, most of the methods simply concatenate the features extracted from eye and face appearance.
The feature fusion process has been ignored. 
In this paper, we propose a novel Adaptive Feature Fusion Network~(\proposed), which performs gaze tracking task in mobile tablets.
We stack two-eye feature maps and utilize Squeeze-and-Excitation layers to adaptively fuse two-eye features according to their similarity on appearance.
Meanwhile, we also propose Adaptive Group Normalization to recalibrate eye features with the guidance of facial feature.
Extensive experiments on both GazeCapture and MPIIFaceGaze datasets demonstrate consistently superior performance of the proposed method.

\end{abstract}


%
\IEEEpeerreviewmaketitle

\section{Introduction}

\renewcommand{\thefootnote}{\fnsymbol{footnote}}\footnotetext[1]{Corresponding author.}
As an important indicator of human attention, gaze is found to be useful to diagnose mental condition and predict human intentions. For example, gaze estimation technology is commonly used in human attention diagnosis like fatigue driving\cite{yoon2019driver,ji2002real} and saliency detection\cite{fan2018salient,wang2017saliency,wang2019inferring}. Gaze has also became a newly-developing human-computer interaction method\cite{zhang2017everyday,piumsomboon2017exploring}, especially in areas like virtual reality\cite{patney2016perceptually,xu2018gaze}.

Up to now, many gaze estimation methods have been proposed. 
Conventional model-based methods estimate gaze by building 3D eye models.
However, they usually require some dedicate devices, like high resolution cameras, RGB-D cameras\cite{alberto2014geometric,xiong2014eye,sun2015real} and infrared cameras\cite{zhu2007novel}.
Recent years, appearance-based methods which directly map gaze from appearance have made great progress. 
Some appearance-based methods using convolutional neural networks (CNNs) have been proposed and show convincing results.
The methods estimating gaze from eye images show reasonable results at the beginning\cite{zhang2015appearance,ranjan2018light}. Later, face images and head pose are found to be helpful for gaze estimation\cite{zhang2017s,Lu2015tip}.
Meanwhile, several large-scale gaze datasets have also been published for the research of CNN-based gaze estimation\cite{huang2017tabletgaze,zhang2017mpiigaze, krafka2016eye}. 
\begin{figure}[htbp]
\centering
\includegraphics[width=\columnwidth]{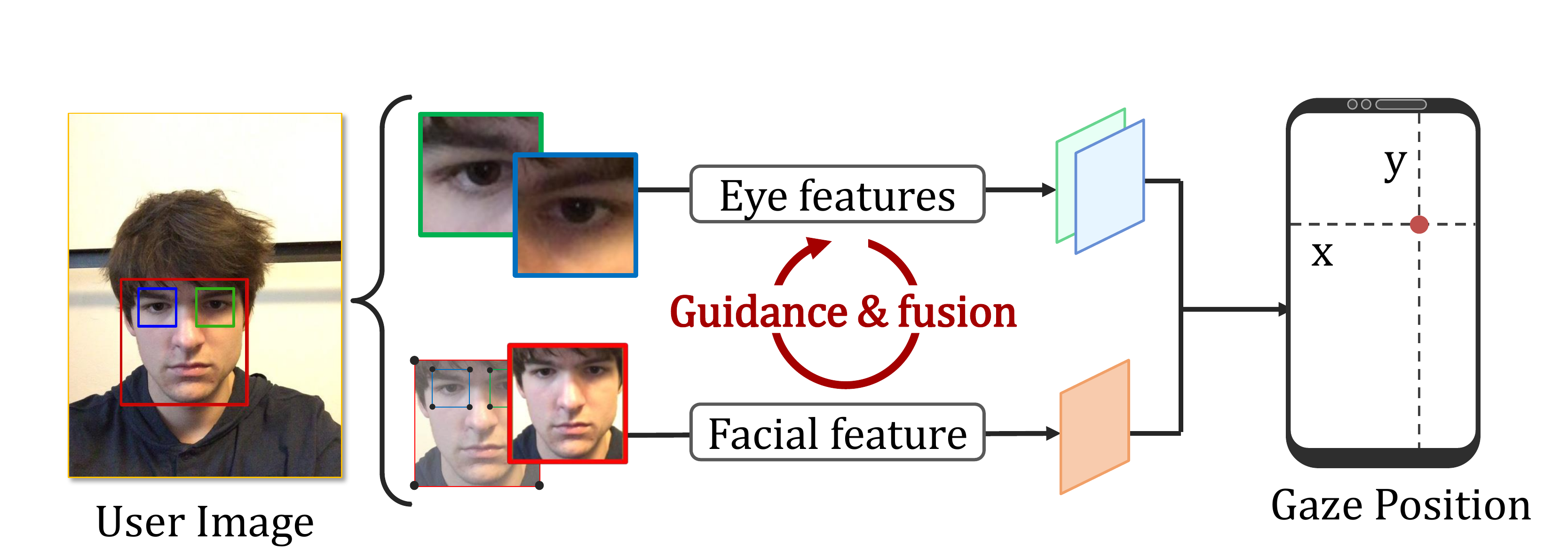} 
\caption{Illustration of our task and proposed technique. User images are used to estimate gaze positions in mobile tablets. An effective guidance \& fusion mechanism is proposed to enhance the feature extraction from both eye regions and facial region.}
\label{intro}
\end{figure}

More recently, CNN-based methods\cite{zhang2017s,krafka2016eye, chen2018appearance} with face and eyes inputs have became popular in gaze estimation since their high accuracy and robustness.
Meanwhile, eye structures like iris and pupil are also crucial for human gaze estimation. Most of CNN-based methods always have eye images~(or face images which contain eyes) for input. Generally, left and right eyes have identical structure and look at the same targets in most time. The observation of obvious relationships between left and right eyes can help to better use eye images for accurate gaze estimation. 
Fischer~\etal~\cite{fischer2018rt} and Krafka~\etal~\cite{krafka2016eye} proposed to concatenate feature vectors from both eyes and process them with fully connected layers (FC layers).
Cheng~\etal~\cite{Cheng2020tip} proposed to utilize two eyes asymmetric by adaptively adjusting evaluation weights of eyes. 
However, we find 1) it's not sufficient to utilize the relationship between both eyes by concatenating feature vectors directly or adjusting weights of eyes. Especially for gaze estimation in wild settings, which is extremely hard to extract and utilize proper eye features.
2) Although face images are found to be helpful\cite{zhang2017s} and become a common input which provide critical information like head pose, light condition, individual differences, 
most methods still treat face and eye images separately. Only few efforts were made to explore face-eye relationship in gaze estimation. Cheng \emph{et al.}\cite{cheng2020coarse} proposed to estimate basic gaze direction from face image and refine it with eye images.

To better utilize the similarity of two eye structures and face-eye relationship, we propose Adaptive Feature Fusion Network (\proposed). We illustrate the pipeline of \proposed~in Figure \ref{intro}. The \proposed~improves gaze tracking accuracy in two ways. First, the \proposed~fuses two eye features according to two eye similarity and appearance. Second, the \proposed~guides eye feature extraction with face appearance characteristics by adaptively recalibrating eye features according to face and eye bounding boxes (we refer to them as Rects) and facial feature. Rigorous experiments show that \proposed~ can produce superior performance against state-of-the-art methods on two commonly used datasets. 
In particular, while combining two eye features, we take two eye feature maps from different layers to stack a fused eye feature map and use a combination of convolutional layers and Squeeze-and-Excitation layers (SE layer) to generate final eye feature. Compared with the classic way of concatenating eye feature vectors, stacked eye feature maps reserve more spatial information and better utilize the identical structure of two eyes. SE layers in the fusion process adaptively weight each channel based on cross-channel information, treat two eye features differently according to their appearance. 
Specifically, 1) to better extract and combine the identical structure of two eyes, we stack a fused eye feature map and use a convolutional layer (conv layer) to generate final eye feature. 
Then we apply SE layer to adaptively weight each channel based on cross-channel information, treat two eye features differently according to their appearance.
2) we also propose a novel Adaptive Group Normalization (AdaGN) to apply face appearance characteristics guidance for eye feature extraction. In detail,  AdaGN takes face, eyes bounding boxes and facial feature as input to scale and shift eye features. 


In summary, the contributions of this paper are as follow:
\begin{itemize}
    \item We propose a novel architecture (\ie, \proposed) for gaze tracking. Motivated by the two eye similarity and relationship between eyes and face, \proposed~equips with better extraction of two eyes' features and face appearance guidance.
	\item We propose a novel Adaptive Group Normalization layer, which recalibrates eye features according to face appearance characteristics. 
	\item The proposed method outperforms existing state-of-the-arts methods on both GazeCapture and MPIIFaceGaze datasets.
\end{itemize}
The rest of the paper is organized as following. Section II summarizes the overview of related works. Section III describes the proposed \proposed. In Section III, we first introduce the general architecture of \proposed. Then, we describe the proposed eye feature fusion scheme and adaptive Group Normalization. Section IV compares the experimental results on two public dataset with several other state-of-the-art methods, conducts ablation studies to evaluate the effectiveness of each component and further analyze the gaze estimation results of proposed \proposed.

\section{Related Works}

As a active research topic, many different approaches have been proposed to address gaze estimation problem. These methods can be categorized into model-based methods and appearance-based methods. 

\textbf{Model-based approaches} aim to fit a 3D eye model to the image and calculate gaze via specific geometric constrains\cite{guestrin2006general}. To acquire accurate eye location, distinct eye features like corneal reflection\cite{nakazawa2012point}, pupil center\cite{valenti2011combining} and iris contour\cite{alberto2014geometric, Lu2016Estimating} are commonly used. Paper\cite{sun2015real} and\cite{xiong2014eye} proposed to estimate gaze using RGB-D cameras. RGB-D cameras are mainly used to obtain the depth of 2D facial landmarks and 3D pupil location in camera coordinate system. Without RGB-D camera, the 3D location of facial landmarks are usually calculated by minimizing projection error between 2D facial landmarks and corresponding points on 3D face model\cite{dementhon1995model}. Wang \emph{et al.}\cite{wang2017real} proposed a deformable eye-face model method. The method models a new subject as a linear combination of offline collected eye-face models. This method also avoid head-eye offset vector estimation to improve accuracy and robustness. Wen \emph{et al.}\cite{Wen2020Eurographics} use convergence constraint which allows calibration without knowing exact gaze location and a person independent gaze corrector to reduce system error. Model-based methods are mostly limited by strict user-camera distance or professional equipment like infrared cameras\cite{nakazawa2012point} and RGB-D cameras\cite{sun2015real,xiong2014eye}. Consequently, model-based methods are accurate under controlled laboratory environment but less reliable under unconstrained environment.

\textbf{Appearance-based approaches} aim to find the direct mapping function from image appearance to gaze direction or gaze location. Appearance-based methods take gaze estimation as a regression from eye images to gaze direction. Thus, they usually only require single camera to capture user face image. Tan \emph{et al.}\cite{tan2002appearance} proposed to estimate gaze with local linear interpolation. Lu \emph{et al.}\cite{lu2014adaptive} proposed an adaptive linear regression, which allows small differences in head rotation, image resolution and blink. Williams \emph{et al.}\cite{williams2006sparse} utilized Gaussian process regression which is able to train on semi-supervised data to estimate gaze direction. Some appearance-based methods directly compute gaze direction based pixel values without training like dimension reduction\cite{lu2017appearance}. Early appearance-based methods usually rely on hand craft features. With dramatic differences of eye appearance under different head orientation, background environment and personal characteristic, it is difficult for hand crafted features to maintain high accuracy in different settings. Thus, the fast developing CNN with strong representational power makes it the most popular method to estimate human gaze.

\textbf{CNN-based methods} attempt to estimate gaze with CNNs, which is trained from several gaze datasets with supervised learning.
Sugano \emph{et al.}\cite{sugano2014learning} capture user images with multiple cameras simultaneously and synthetic images with different gaze directions to compose the UT Multiview dataset. Funes \emph{et al.}\cite{funes2014eyediap} published the Eyediap dataset. 16 subjects were required to look at both 2D screen targets and 3D floating targets. Zhang \emph{et al.}\cite{zhang2017mpiigaze} proposed a CNN with facial weights and published the popular MPIIGaze dataset. Zhu \emph{et al.}\cite{zhu2017monocular} process head pose information with gaze transfer layer to increase robustness by eliminate the overfitting of head-gaze correlation that differs for every dataset. Kim \emph{et al.}\cite{kim2019nvgaze} published near infrared dataset NVGaze, which contains big amount of collected real images and high resolution synthetic images with detailed parameters. Cheng \emph{et al.}\cite{cheng2020coarse} proposed a coarse-to-fine way which estimates a basic gaze direction from face images and refine it with more detailed eye images. To address the lack of large scale dataset, Krafka \emph{et al.}\cite{krafka2016eye} collected the GazeCapture dataset with mobile devices and proposed iTracker which takes eyes, face and face grid as input. Based on the GazeCapture dataset, gaze estimators under mobile devices settings gains much attention. He \emph{et al.}\cite{he2019device} proposed a lite CNN model without face input and a few shot calibration scheme which further increases accuracy. Guo \emph{et al}\cite{guo2019generalized} proposed a Tolerant and Talented training scheme to get better accuracy and robustness. Thanks to large dataset like GazeCapture and powerful feature extraction ability of CNN, appearance-based methods achieve good accuracy and head-pose independence. The ability to tolerant low resolution images and changing environments also make appearance-based methods more effective in wild settings.

\section{Adaptive Feature Fusion Network}

\begin{figure*}[!t]
\centering
\includegraphics[width=2\columnwidth]{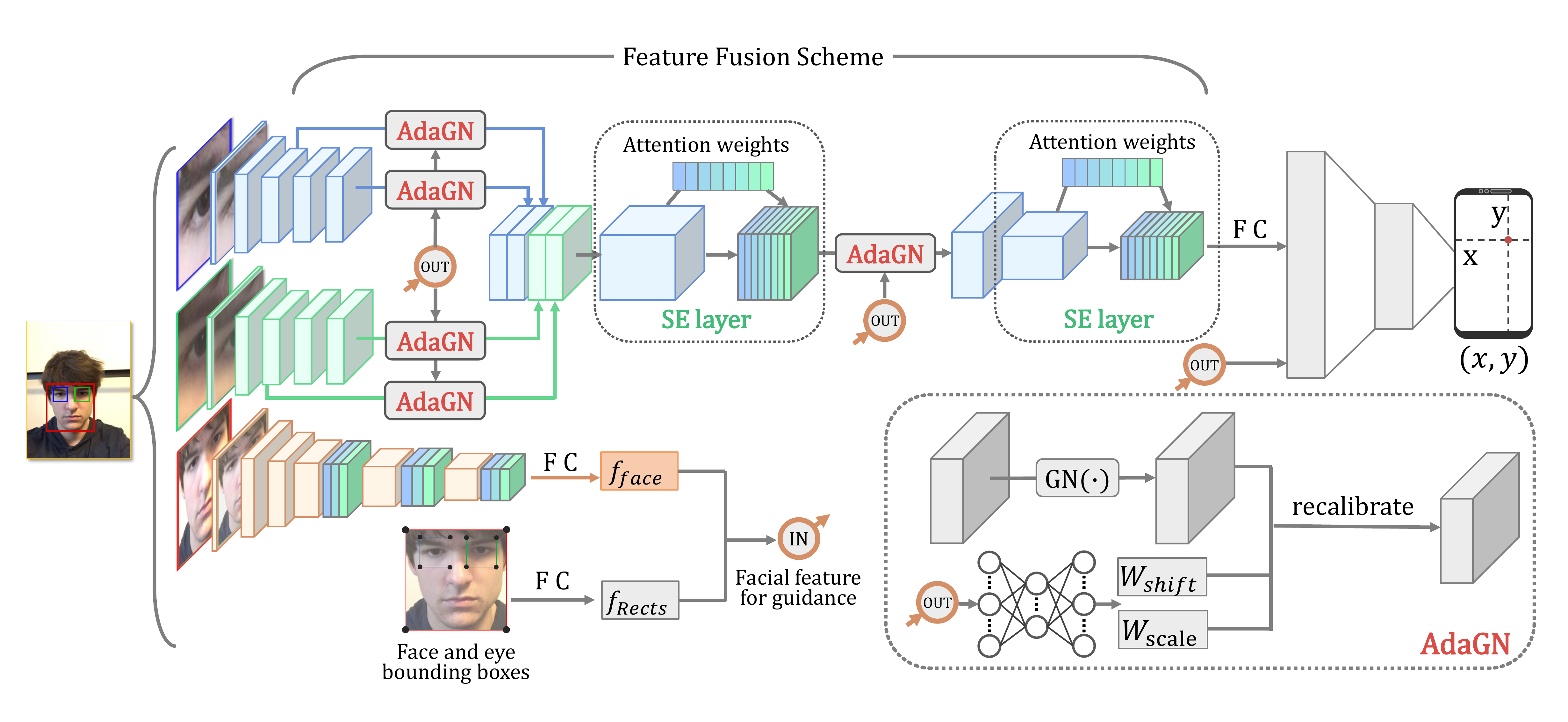} 
\caption{Structure of proposed \proposed. Facial feature is extracted by several conv layers and FC layers. Rects feature is extracted directly by FC layers. The eye stream takes left and flipped right eye images as input and extract fused eye features by proposed stacking architecture and SE layers. AdaGN recalibrates eye features with face appearance characteristics derived from face and Rects features.}
\label{summary}
\end{figure*}

\subsection{Overview of the network}
The structure of \proposed~is shown in Figure \ref{summary}. The \proposed~has four inputs, which are face image, left and right eye images,  top left corner and bottom right corner of face and eye bounding boxes . We propose \proposed~in order to improve eye tracking accuracy by 1) utilizing the similarity of two eyes appearance to adaptively fuse eye features, 2) guide eye feature extraction with face appearance characteristics. For the first aspect, we stack feature maps from both eyes in channel wise to fuse eye features, which are followed by several convolutional layers. SE layers are used during eye feature extraction to weight features from different eyes according to their appearance. For the second aspect, the pre-processed Rects and facial feature are normalized for guiding eye feature extraction.
\subsection{Squeeze and excitation for eye feature fusion}
When fuse eye features, most state-of-the-arts methods concatenate eye feature vectors reshaped from eye feature maps and process them with fully connected layers. However, there are two problems in these methods. First, reshape feature maps to feature vector loses some spacial and cross-channel information, weakening two eyes relationships. Second, fully connected layers are not as powerful as convolutional layers when dealing with images. To handle such problems, we propose an eye feature fusion structure which contains eye feature stacking and SE layer for adaptive eye fusion. 
\begin{figure}[tbp]
\centering
\includegraphics[width=\columnwidth]{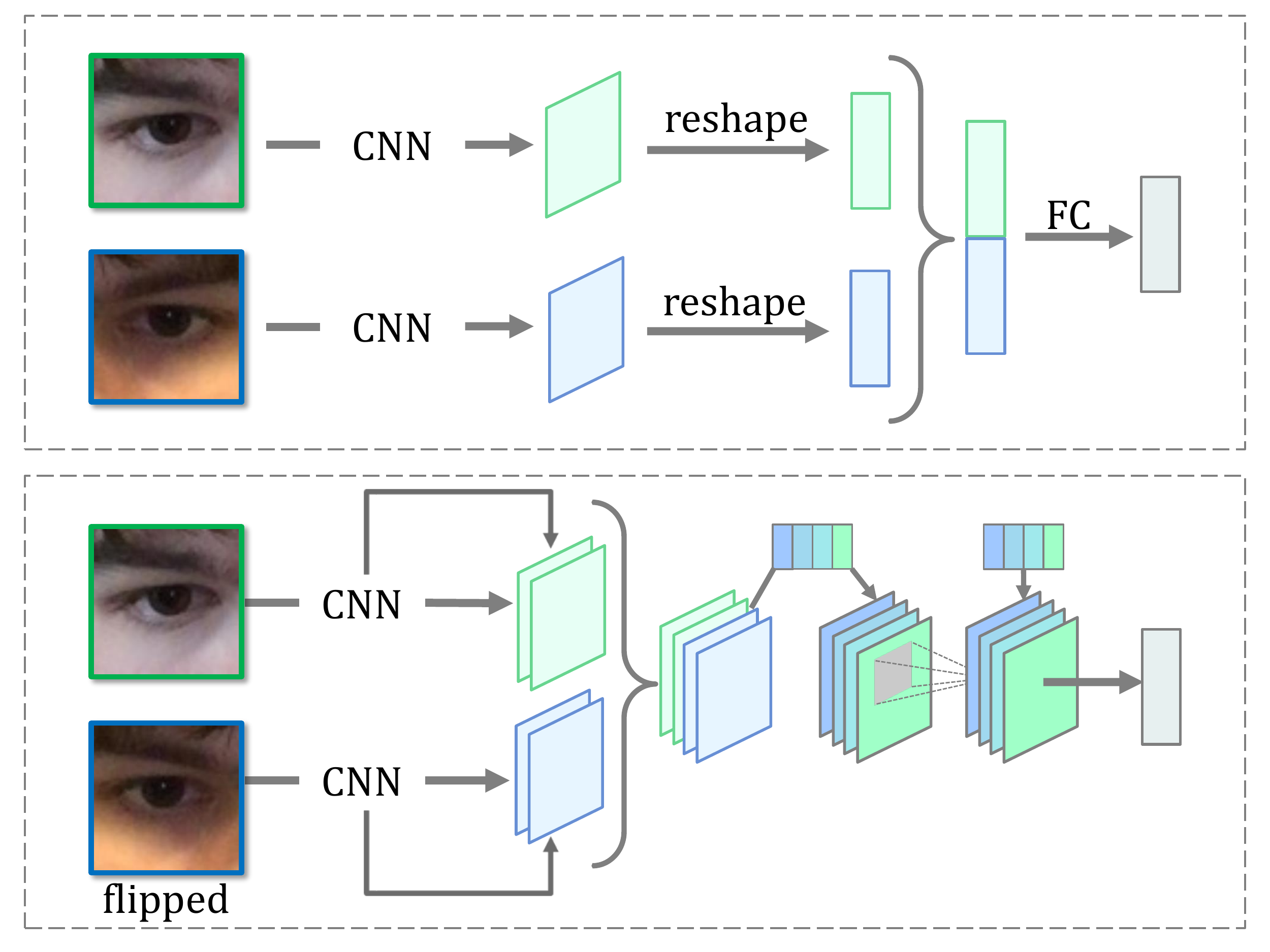} 
\caption{Classic eye feature fusion structure (top) and proposed eye feature fusion structure (bottom).}
\label{EyeFusion}
\end{figure}
As shown in Figure \ref{EyeFusion}, we take eye feature maps from different layers and stack them in channel wise. For the shape and structures of two eyes are identical, it is reasonable to fuse eye features by stacking feature maps together. We take feature maps from lower layer for they reserve more spacial information and feature maps from higher layer for their stronger representational abilities. The fused eye features are followed by convolutional layers to generate final eye feature vector. In this way, two eyes relationships are considers by calculating the final eye feature values according to the corresponding areas in both eyes.
	
SE layer is a powerful structure for applying attention for different eye features from different channels. SE layer works as Equation \ref{EquaSE}:
\begin{gather}
\begin{cases}
W_{weight}=\sigma(L(GAP(f_{in}))),\\ \label{EquaSE}
f_{out}=F_{scale}(W_{weight},f_{in})),\\
\end{cases}
\end{gather}
$f_{in}$ stands for the input feature of SE layer. $GAP(\cdot)$ stands for Global Average Pooling layer (GAP layer). Every channel of input feature is compressed to single value by GAP layer. $L(\cdot)$ stands for FC layers and $\sigma{\cdot}$ stands for Sigmoid function. Then, a weight vector $W_{weight}$ is calculated by FC layers. $F_{scale}(\cdot)$ stands for channel wise multiplication of inputs. $W_{weight}$ is applied to original input to derive final result $f_{out}$. SE layer allocates attentions to different channels according to cross-channel relationships, which is suitable for two-eye relationship based eye feature fusion.

In the eye feature fusion structure, SE layers are added before and after eye feature stacking.
Before eye feature stacking, SE layers are added to dynamically adjust channel-wise features, enhance the representational power of the network\cite{hu2018squeeze}. After eye feature stacking, as described above, spacial information and high level features from left and right eyes are stored in different channels of fused feature map. To summary, the SE layers are used for adaptively balancing spatial information and complex features from left and right eye according to two eye appearance (eye feature map values) and relationship (cross-channel information).

Note that right eye images are horizontally flipped in data processing. As we stack two eye features in channel wise, it is vital to ensure the extraction of two eye features are identical. To address such issue, we 1) use shared weights CNN to process eye images, and 2) ensure the appearance of two eyes stays the same by flipping. Through flipping, the position of inner and outer eye corners, the direction of eyebrows are consistent in left and right eye images due to the different eye orientation. This makes it easier for shared weights Eye Net to extract eye features like iris for the different appearance, further improves the final performance.
\subsection{Adaptive Group Normalization}
To fuse facial feature for a better gaze estimation, we propose the Adaptive Group Normalization (AdaGN) that adequately utilizes facial feature guidance.
Though eye features are vital for gaze estimation, it is also difficult to extract proper eye features for eye appearance changes dramatically due to different factors like head pose, light condition and individual differences. The small region of the eye makes it difficult for the network to recognize all these changing factors. On the contrary, these factors always result in the difference of face location and appearance. Thus, it is necessary to consider face appearance characteristics during eye feature extraction procedure. In CNNs, some normalization layers contains scale and shift operation to enhance the representational power of normalized features. Inspired by them, we guide recalibration in Group Normalization with face appearance characteristics. We propose AdaGN which takes concatenated Rects features and facial features as input to represent face appearance characteristics. Then, AdaGN adaptively adjusts eye feature extraction according to face appearance characteristics by recalibrating eye features. According to different combination of head pose, light condition and other factors reflected at face appearance characteristics, AdaGN calculate scale and shift parameters to adaptively recalbrate eye features.

In detail, we first put Rects through four fully connected layers to generate a 64 dimensional vector which contains information about head translation. Then, the 64 dimensional feature vector is concatenated with facial feature to represent face appearance characteristics. The AdaGN works as follow equations:
\begin{gather}
\begin{cases}
[W_{shift}, W_{scale}]=LeakyReLU(L(f_{Rects}, f_{face})),\\ \label{EquaAdaGN}
f_{out}=W_{scale}*GN(f_{in})+W_{shift},\\
\end{cases}
\end{gather}
where $L(\cdot)$ stands for fully connected layer, $GN(\cdot)$ stands for normal Group Normalization without scale and shift. $f_{in}$ is the original feature map and $f_{out}$ is the final output of AdaGN. Rects features extracted by FC layers and facial feature extracted by face stream are represented as $f_{Rects}$ and $f_{face}$ in Equation \ref{EquaAdaGN}. $W_{scale}$ and $W_{shift}$ are scale and shift parameters for each channel of $f_{in}$. AdaGN in eye net takes the Rects features and facial feature as input and calculate shift and scale by a fully connected layer. We choose Group Normalization rather than Batch Normalization for BN normalizes every channel in batch wise and neglects cross-channel relationship, adding disturbance to the channel weights given by SE layers. In this way, \proposed~extracts eye features adaptively based on face appearance characteristics, generates more reasonable eye features for accurate gaze tracking. 

\subsection{Point of gaze prediction}
In this section, we describe the complete architecture of proposed \proposed. For eye streams, feature maps from third and fifth convolutional layers are stacked to forge fused eye feature map. Then, another conv layer produces the final eye feature. AdaGNs are used as normalization layers. For face stream, facial feature is extracted by 6 conv layers. Last three conv layers are followed by SE layers to enhance representational power. For Rects stream, 64 dimensional feature vector generated by four FC layers is used as both the input of AdaGN and final Rects feature.
Eye, face and Rects features are concatenated and fed to two FC layers to predict 2D gaze position of screen. We use ReLU as excitation layers for all convolutional layers and LeakyReLU for all FC layers. Similar to\cite{krafka2016eye}, we use physical distance related to the camera as label to achieve device independence. Adam optimizer and Smooth L1 Loss function are used while training.

\begin{table}[h]
\centering
\renewcommand\arraystretch{1.5}
\caption{Gaze estimation error in centimeters compares with SOTA methods on the GazeCapture dataset.}
\begin{tabular}{p{0.2\columnwidth}<{\centering}|p{0.15\columnwidth}<{\centering}p{0.15\columnwidth}<{\centering}}
\hline
Method      & Phone error (cm)          & Tablet error (cm)          \\ \hline \hline
iTracker\cite{krafka2016eye}  & 1.86  &  2.81    \\ \hline 
SAGE\cite{he2019device}          &    1.78        &   2.72        \\ \hline
TAT\cite{guo2019generalized}   &    1.77        &   2.66          \\ \hline \hline
\proposed                   &   \textbf{1.62}  & \textbf{2.30}          \\ \hline
\end{tabular}
\label{comparisontable}
\end{table}
\begin{figure}[h]
\centering
\includegraphics[width=\columnwidth]{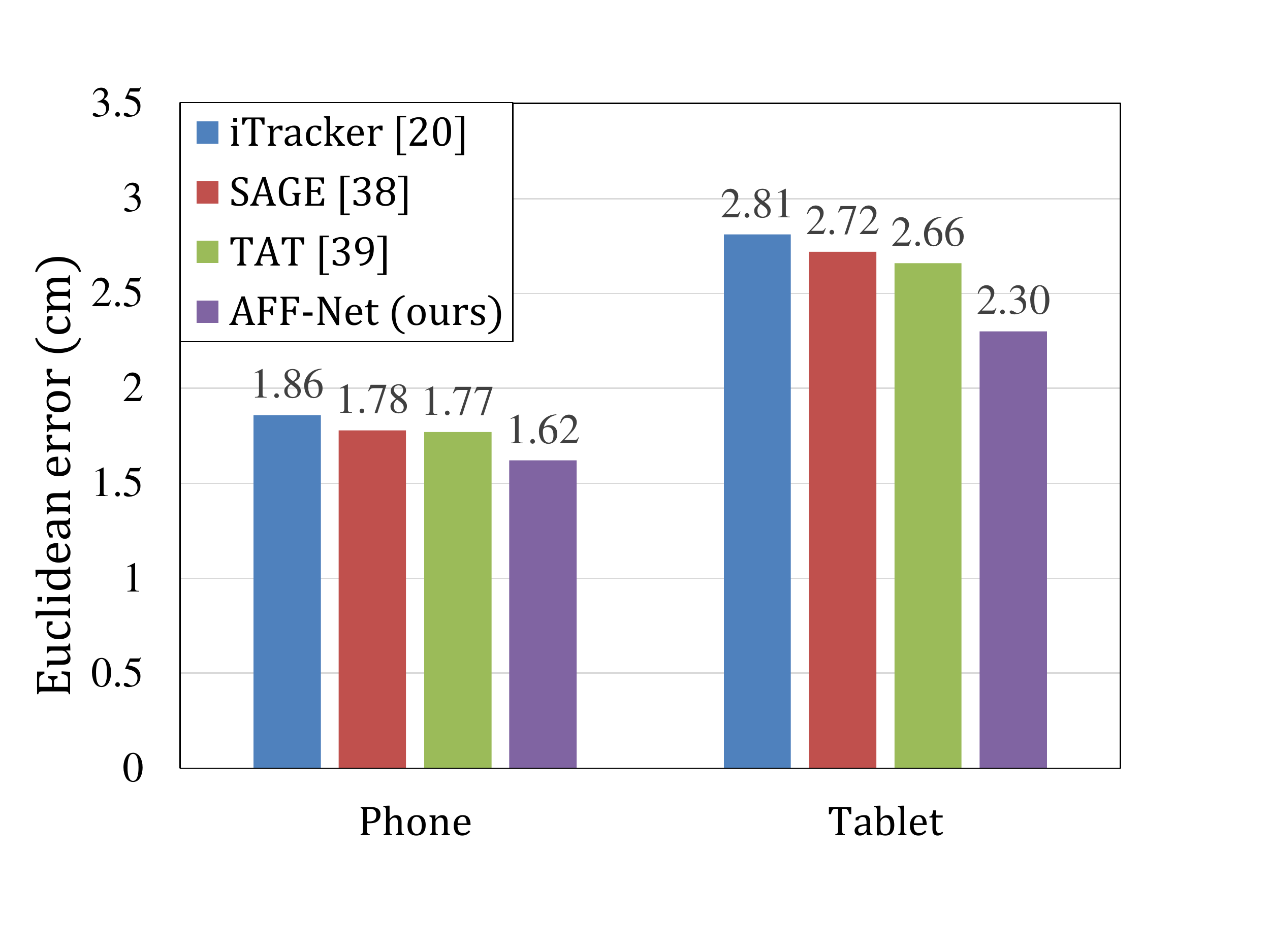} 
\caption{Gaze estimation error in centimeters compared with SOTA methods on the GazeCapture dataset.}
\label{PerformanceComparison}
\end{figure}

\section{Experiments}
\subsection{Setup}
\subsubsection{Dataset}
We conduct experiments in two famous dataset with 2D gaze labels: GazeCapture\cite{krafka2016eye} and MPIIFaceGaze\cite{zhang2015appearance}.

GazeCapture is a large 2D gaze dataset with 2,445,504 images from $\sim$1,500 subjects. The GazeCapture dataset is collected using crowdsourcing. The dataset is captured by mobile phones or tablets in different orientations. There are 1,490,959 frames have both face and eye detections, which are further divided into 1,251,983 training images, 59,480 validation images and 179,496 test images. GazeCapture provides gaze positions represented as pixel location on screen and physical distance to the camera. We use physical distance to the camera in both training and testing for its device independence.

MPIIFaceGaze contains 213,659 images from 15 subjects. It is a commonly used dataset for gaze estimation problem. MPIIFaceGaze has a larger prediction space than GazeCapture for it is collected by laptops. The dataset provides physical screen sizes and 2D gaze position in pixels. So we convert pixel coordinates of the gaze to physical coordinates relative to the screen.

\subsubsection{Data processing}
For the GazeCapture dataset, we directly crop face and eye images according to face detection results of python face-recognition lib. Face images are resized to 224*224*3, and eye images are resized to 112*112*3. Pixel values are normalized into [0, 1]. For the MPIIFaceGaze dataset, we calculate face and eye bounding boxes from 6 facial landmarks in the dataset label. 
Specifically, we take 1.7 times eye x coordinates width as eye bounding box size and the average of eye coordinates as eye center. We set the face center as the mid point of the average eye coordinates and the average mouth coordinates. Face bounding box size is $1/0.3$ times eye bounding box size.
We use the same image resolution settings as in the GazeCapture dataset.
\begin{table}[t]
\centering
\renewcommand\arraystretch{1.5}
\caption{2D gaze position estimation error in centimeters and 3D gaze direction estimation error in degrees compared with other major methods on the MPIIFaceGaze dataset.}
\begin{tabular}{p{0.2\columnwidth}<{\centering}|p{0.2\columnwidth}<{\centering}p{0.2\columnwidth}<{\centering}}
\hline
Method      & 2D gaze location error (cm)          & 3D gaze direction error (degree)          \\ \hline \hline
iTracker\cite{krafka2016eye}  & 5.46  &  6.2    \\ \hline 
Spatial weights CNN\cite{zhang2017s}    &    4.2    &   4.8       \\ \hline
RT-GENE\cite{fischer2018rt}   &   4.2        &   4.8         \\ \hline \hline
\proposed                   &   \textbf{3.9}  & \textbf{4.4}          \\ \hline
\end{tabular}
\label{comparisontable_mp}
\end{table}
\begin{figure}[t]
\centering
\includegraphics[width=\columnwidth]{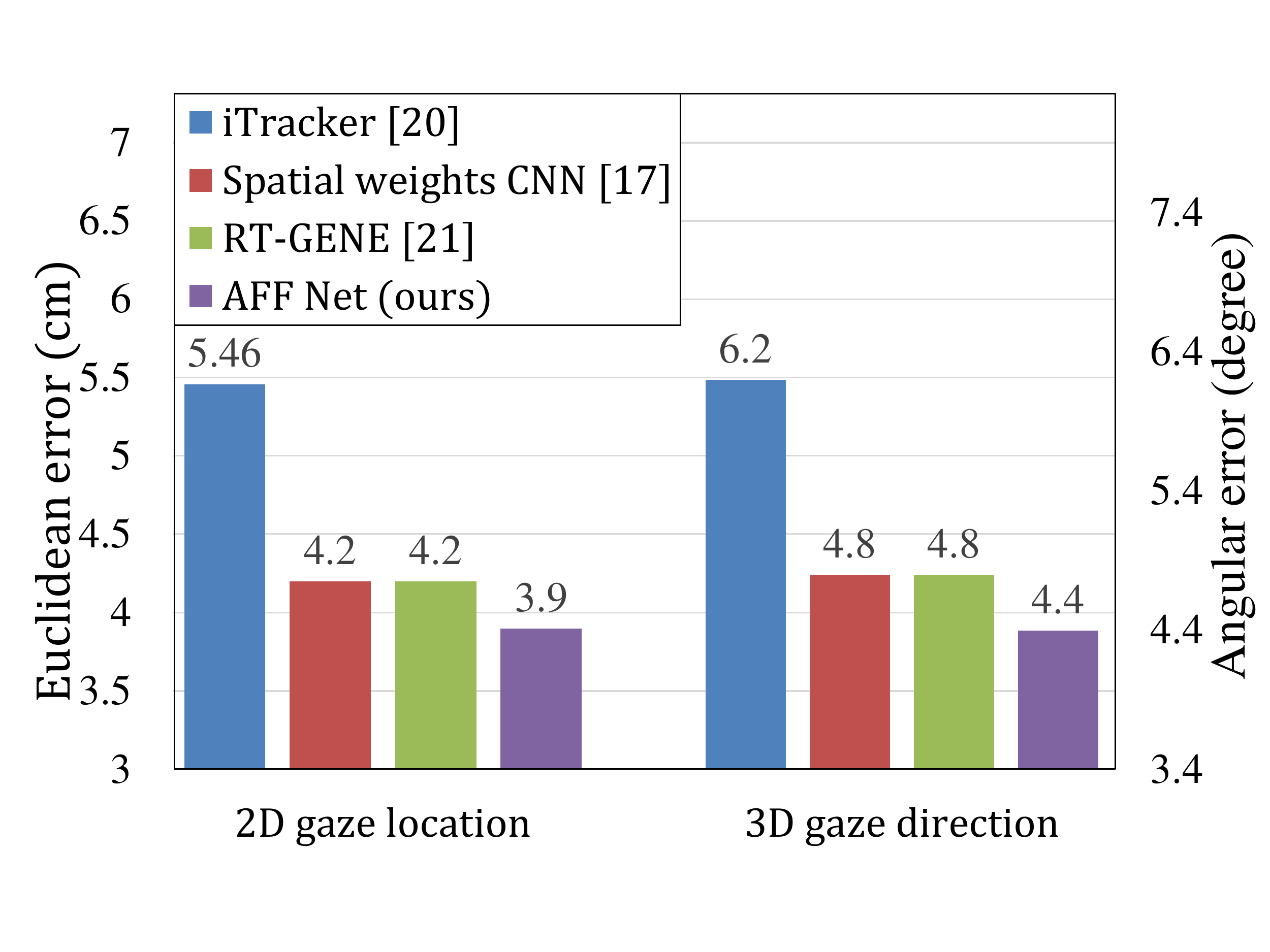} 
\caption{2D gaze position estimation error and 3D gaze direction estimation error compared with other major methods on the MPIIFaceGaze dataset.}
\label{PerformanceComparison_mp}
\end{figure}
In GazeCapture, we follow the given train and test set. To simulate settings without calibration, we do leave-one-person-out test and average results from all subjects as final performance in MPIIFaceGaze.

\subsubsection{Implementation details}
The face stream has 6 conv layers. The specific parameters represented as (out channels, kernel size, stride, padding) are [(48, 5, 2, 0), (96, 5, 1, 0), (128, 5, 1, 2), (192, 3, 1, 1), (128, 3, 2, 0), (64, 3, 2, 0)]. 3x3 max pooling layers with stride 2 is used after the second and third conv layers. Two FC layer further compress the facial feature to a 64 dimensional vector. Rects input is processed by four FC layers with output channels as (64, 96, 128, 64).

The eye stream has 5 conv layers with parameters as [(24, 5, 2, 0), (48, 5, 1, 0), (64, 5, 1, 1), (128, 3, 1, 1), (64, 3, 1, 1)]. Max pooling layers in eye stream is the same as in face stream. 
After feature maps from third and fifth conv layers are fused, a (64, 3, 2, 1) conv extracts the joint two eyes feature map. One FC layer compress the two eyes feature map to a 128 dimensional vector. The final eye, face and Rects feature vectors are concatenated and two FC layers with 128, 2 dimensional outputs derive the gaze position coordinates. For the left and right eye networks share weights, right eye image is flipped to keep the eye orientation consistent. SE layers are added after second, forth, last conv layers and stacking.

The model is trained for 12 epochs on the GazeCapture dataset with Smooth L1 Loss function. The learning rate is 0.001 and reduce to 0.0001 after 8 epochs. Batch size is set to 256. On the MPIIFaceGaze dataset, the model is trained for 17 epochs under same settings. Similar to \cite{krafka2016eye,he2019device}, we add a random shift from 0 to 30 pixels to the face and eye positions while training to improve the robustness of the model. For testing, we report Euclidean distance between prediction and ground truth in centimeters. The whole experiments are implemented using PyTorch. Network parameters are initialized by the default initialization of PyTorch.

\subsection{Performance}
The GazeCapture dataset is almost the largest gaze dataset in mobile device.
We first conduct performance evaluation of our method on the GazeCapture dataset.
We choose three methods for comparison on GazeCapture, which are SAGE~\cite{he2019device}, TAT~\cite{guo2019generalized} and iTracker~\cite{krafka2016eye}. 
To the best of our knowledge, TAT shows the state-of-the-art performance on GazeCapture.
We illustrate the result in Figure~\ref{PerformanceComparison} and list the result in Table~\ref{comparisontable}. 
Our \proposed~achieves 1.62 cm error on mobile phone captured images and 2.30 cm error on tablet captured images, outperforms state-of-the-art methods. For mobile phone image test, the earliest iTracker has the highest error as 1.86 cm. SAGE and TAT has similar performance around 1.77 cm, improve about 5\% from iTracker. Our \proposed~achieves 1.62 cm error, outperforms other methods significantly. The \proposed~improves about 8.5\% from SAGE and TAT. For the more challenging tablet image test, the error of iTracker is 2.81 cm. Different from mobile phone image test, TAT achieves 2.66 cm error, which is 0.06 cm lower than SAGE. Our \proposed~also has the lowest error at 2.30 cm, which significantly improves 13.5\% from the second best method TAT. As there are only about 15\% images are from tablets, the results show that \proposed~learns proper features faster than other methods. In the mean time, our \proposed~only has 1.94M parameters. It is 3 times fewer compares to the iTracker which has 6.29M parameters. These experiment results show that \proposed~has a clear advantage compare with other methods, especially in tablet images which is less in total amount.


To further demonstrate the advantage of our method, We conduct more experiments on the MPIIFaceGaze dataset.
We calculate face and eye bounding boxes according to provided facial landmarks and convert the screen pixel coordinates of targets to physical distance.
We choose iTracker~\cite{krafka2016eye} and Spacial Weights CNN~\cite{zhang2017s} as the compared methods, since they both show outstanding performance in the 2D gaze position estimation task on MPIIFaceGaze.
Meanwhile, since MPIIFaceGaze is popular used in 3D gaze direction estimation task,
we also select RT-GENE~\cite{fischer2018rt} as compared method for providing convinced comparison. The RT-GENE almost shows start-of-the-art performance in 3D gaze direction estimation task on MPIIFaceGaze.
In order to provide more comprehensive comparison, we further convert the 2D gaze positions result estimated from our \proposed~into 3D gaze directions according to provided camera-screen calibration matrix.
As can be seen in Figure \ref{PerformanceComparison_mp} and Table \ref{comparisontable_mp}, our method achieves the performance of 3.9 cm Euclidean error and 4.4 degree angular error, which significant performs better than other compared methods. 
Note that, the MPIIFaceGaze dataset is collected from laptop.
This result demonstrate that our method also can perform well in the laptop.


\subsection{Ablation Study}
In this section, we conduct ablation experiments to prove the advantage of each proposed method. We focus on three component described above: stacking (\emph{ST}), SE layers for eye feature fusion (\emph{SE}) and adaptive Group Normalization (\emph{AdaGN}).
\begin{table}[]
\centering
\renewcommand\arraystretch{1.5}
\caption{Euclidean error in centimeters on the GazeCapture dataset for ablation study.}
\begin{tabular}{p{0.25\columnwidth}<{\centering}|p{0.15\columnwidth}<{\centering}p{0.15\columnwidth}<{\centering}}
\hline
\multirow{2}{*}{Method} & \multicolumn{2}{c}{GazeCapture} \\ \cline{2-3} 
                        & Phone (cm)          & Tablet (cm)          \\ \hline \hline
\proposed                 & \textbf{1.62}  &   \textbf{2.30}   \\ \hline \hline
without ST                &    1.67        &   2.39        \\ \hline
without SE                &    1.68        &   2.31          \\ \hline
without AdaGN                   &    1.69        &   2.33          \\ \hline
\end{tabular}

\label{ablationtable}
\end{table}
Specifically, to ablate SE layers, we simply remove all SE layers in the \proposed~and keep other components the same. This architecture is denoted as \emph{No SE} in \ref{ablationtable}. To ablate AdaGN, the normal GN without extra input is used instead of AdaGN with Rects input. When ablates stacking, we carefully ensure the number of conv layers keeps the same for fair comparison. We remove the stacking architecture and directly move the conv layer after stacking to the end of eye stream. Then, the feature maps from left and right eyes are reshaped to feature vectors and processed by FC layers like classic methods. 

The results of ablation study is shown in Table \ref{ablationtable}. Complete \proposed~with all three components achieves the highest accuracy. Removing of stacking architecture increases error on both phone images and tablet images by 0.05 cm and 0.09 cm respectively. Ablation of AdaGN causes a 0.07 cm error increase on phone images. Model without SE layers also has a 0.06 cm error increase on phone images, while the accuracy on tablet images remains nearly the same. The results show that removing any one of three components will increase the Euclidean error by about 4\%, proves the effect of proposed stacking architecture, SE layers and AdaGN.
\begin{figure}[t]
\centering
\includegraphics[width=\columnwidth]{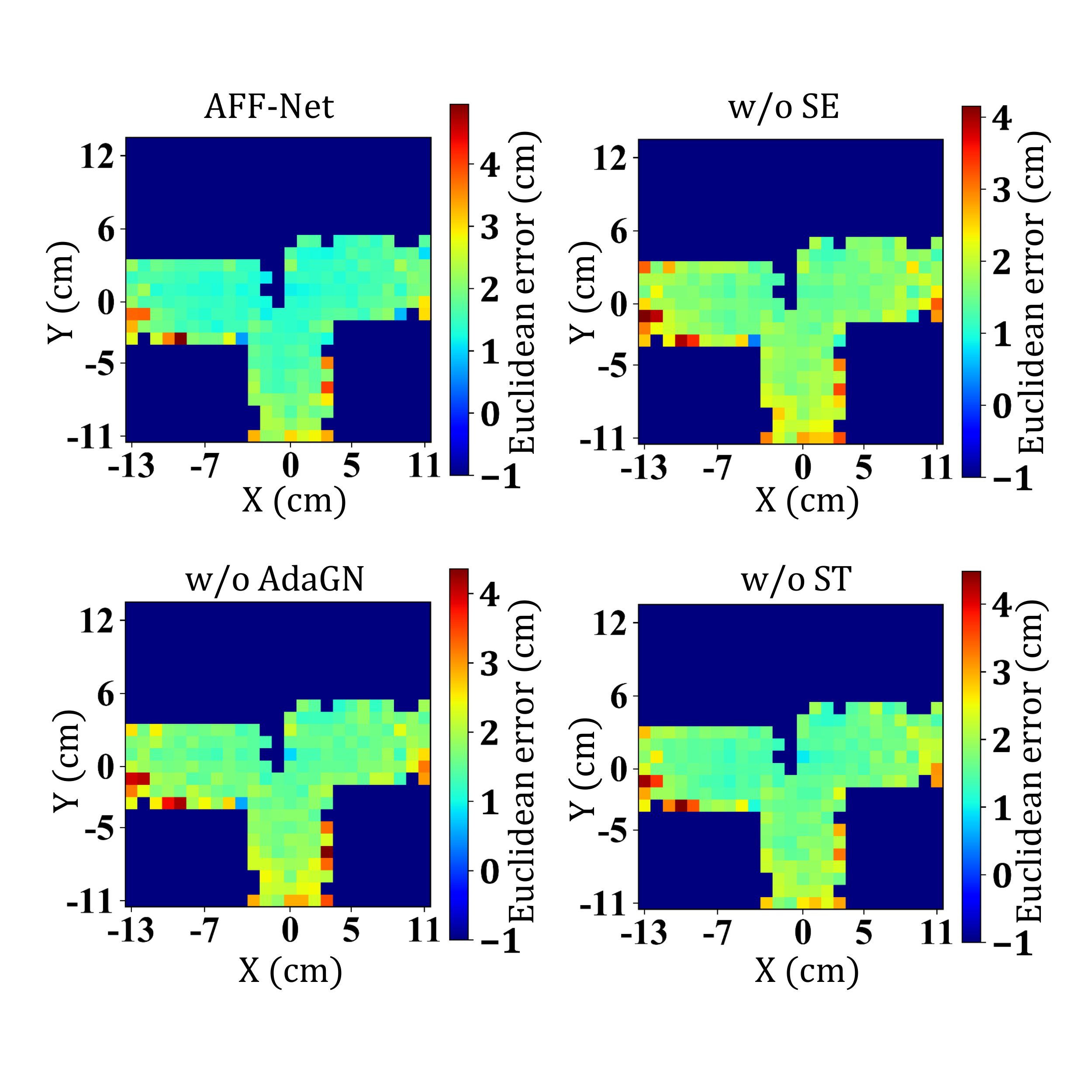} 
\caption{Spacial heat maps of Euclidean error in centimeters on the GazeCapture dataset for different models.}
\label{heatMap}
\end{figure}
\subsection{Result Analysis}
In this section, we further analyze the result of different models presented in ablation study on phone images of the GazeCapture dataset. Figure \ref{heatMap} shows heat maps of gaze estimation error in centimeters for different models. The camera is at the (0, 0) in every heat map. Gaze locations result in only three branches for GazeCapture does not contain phone test images with upside down orientation. As can be seen for all four heat maps, gaze estimation error increases as gaze location moves away from camera. The error of \proposed~clearly increases slower than other three ablated networks, proving stronger robustness for different gaze location.

\begin{figure}[t]
\centering
\includegraphics[width=\columnwidth]{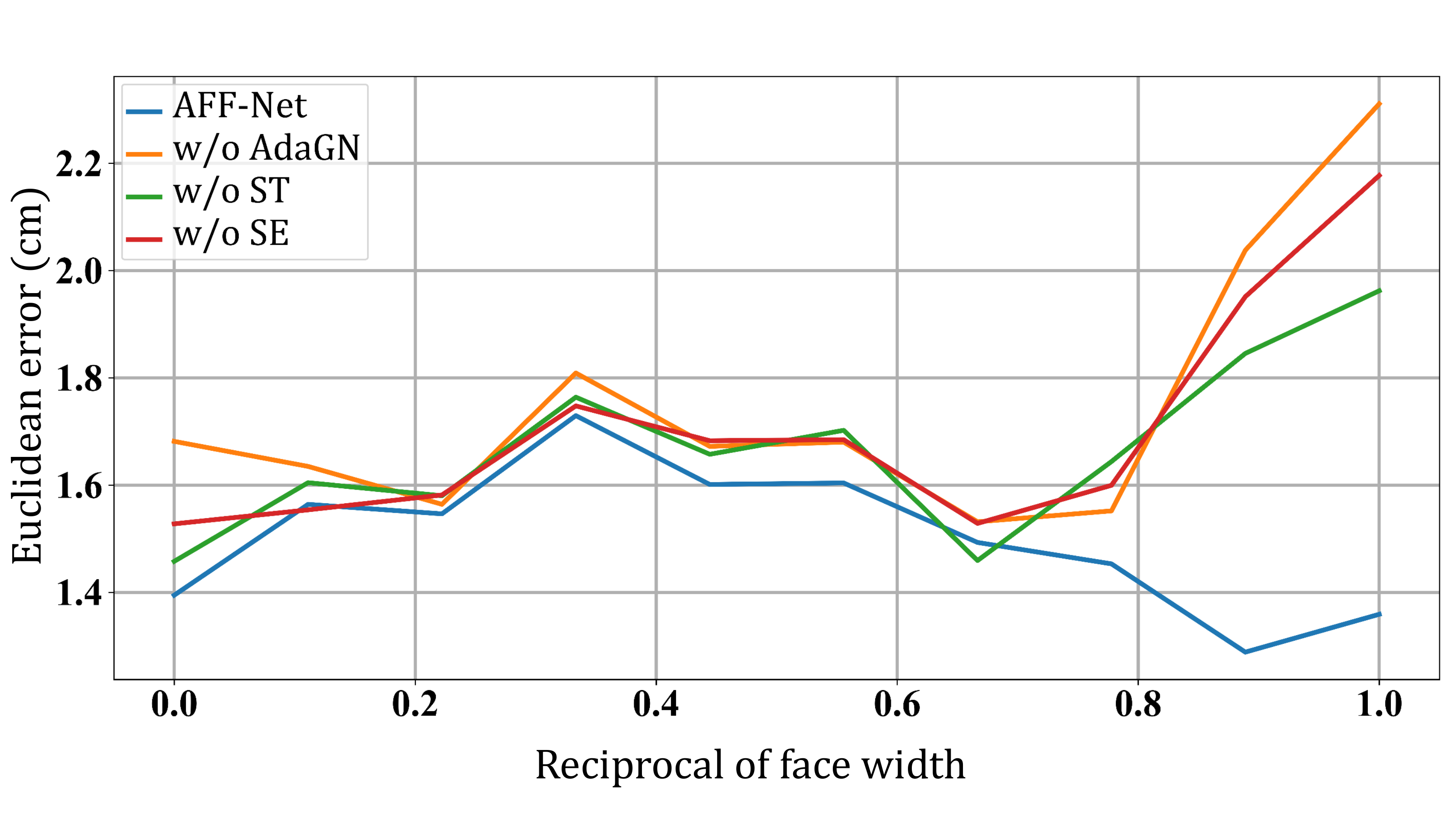} 
\caption{Curve of Euclidean error in centimeters on GazeCapture dataset relative to the reciprocal of face width.}
\label{facedis}
\end{figure}

Figure \ref{facedis} shows the curve of gaze estimation error relative to reciprocal of face width. Specifically, we set $X$ axis as the reciprocal of face width relative to the shorter axis of original image. We choose it as $X$ axis for it generally reflects the user-camera distance as face appears smaller when user moves away from camera, although it is disturbed by camera length and individual differences. As shown in Figure \ref{facedis}, the \proposed~evidently outperforms other models, especially in images with very small face size. This results show the better robustness of \proposed~against extreme cases.

\section{Conclusion}
In this paper, we propose an accurate appearance-based gaze estimation method named \proposed. The proposed \proposed~improves gaze tracking accuracy by adaptively fusing two eye features and face appearance characteristics guided eye feature extraction. In particular, we propose a stacking architecture with SE layers to fuse two eye features. Inspired by the identical structure of two eyes, we stack feature maps from different eyes and derive final feature map by convolutional layers. SE layers are added in the fusion process to adaptively weight two eye features according to their appearance. In addition, we also propose to calculate shift and scale parameters in Group Normalization based on face appearance characteristics to realize eye feature recalibration. The \proposed~which combines above methods achieves state-of-the-art performance on both GazeCapture and MPIIFaceGaze dataset, proves the effectiveness of proposed network.

%




\bibliographystyle{IEEEtran}
\bibliography{ref}

\end{document}